# Image Deblurring using GAN

**Zhengdong Li**

Department of Electronic and Computer Engineering

HKUST

zlifd@connect.ust.hk

**Abstract**

In recent years, deep generative models, such as Generative Adversarial Network (GAN), has grabbed significant attention in the field of computer vision. This project focuses on the application of GAN in image deblurring with the aim of generating clearer images from blurry inputs caused by factors such as motion blur. However, traditional image restoration techniques have limitations in handling complex blurring patterns. Hence, a GAN-based framework is proposed as a solution to generate high-quality deblurred images. The project defines a GAN model in Tensorflow and trains it with GoPRO dataset. The Generator will intake blur images directly to create fake images to convince the Discriminator which will receive clear images at the same time and distinguish between the real image and the fake image. After obtaining the trained parameters, the model was used to deblur motion-blur images taken in daily life as well as testing set for validation. The result shows that the pretrained network of GAN can obtain sharper pixels in image, achieving an average of 29.3 Peak Signal-to-Noise Ratio (PSNR) and 0.72 Structural Similarity Assessment (SSIM). This help to effectively address the challenges posed by image blurring, leading to the generation of visually pleasing and sharp images. By exploiting the adversarial learning framework, the proposed approach enhances the potential for real-world applications in image restoration.

## 1    Introduction

Generative adversarial networks (GANs), introduced by Ian J. Goodfellow [1] in 2014, is unsupervised deep learning machine . It involves 2 components: the Generator and the Discriminator. The Generator tries to mimic a target data distribution while the Discriminator tries to distinguish the samples coming from the Generator and the real data. In the recent years, with the rising concepts of deep learning, multiple methods based on GAN have been proposed for image deblurring task. For example, [2] proposed a learning method based on GAN and content loss to construct the Generator and Discriminator. Blind recovery of blurred images is achieved via adversarial competition between the two networks. [3] proposed a blind deblurring for sharp image reconstruction using dense neural network. However, both [2] and [3] do not involve a detail discussion on the network architecture, such as the number of layers and parameters for Generator and Discriminator. [4] proposed an intensive GAN by increasing the number of residual modules in the network, achieving a better improvement in image recovery quality. Yet, key factors like the number of parameter is also missing. Hence, this project is going to fill in the missing gap for these, involving a comprehensive discussion on image deblurring using GAN.

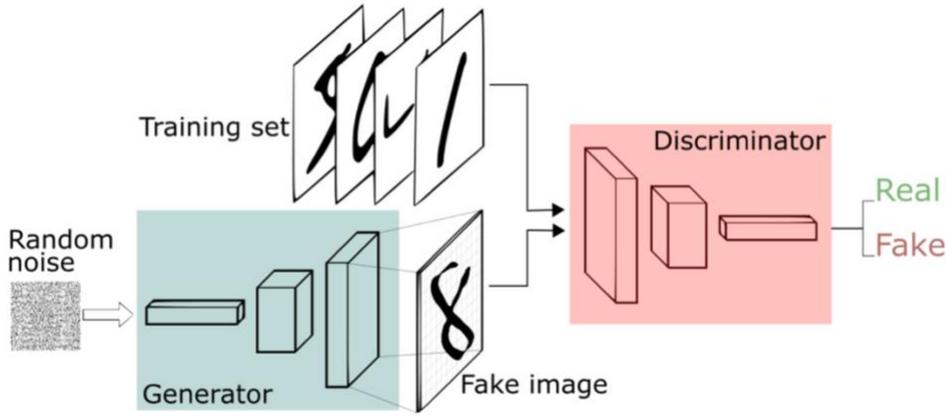

Figure 1: The GAN architecture for image deblurring [5]

Figure 1 refers to the working principle of image deblurring using GAN. The Generator takes random noise as input and produces a fake image that resembles a real image. Its goal is to fool the Discriminator into believing that the fake image is the real one. On the other hand, the Discriminator is presented with both real and fake images and aims to accurately distinguish between them. It avoids to fall into the trap given by the Generator and strives to output "True" for real images and "False" for fake ones. This feedback helps to guide the Generator in improving its output to create more convincing images that deceive the Discriminator. In the adversarial game between the Generator $G$ and the Discriminator $D$, both parties have opposing objectives. With the addition of random noise $z$ and real data $x$, the Discriminator aims to maximize its values, while the Generator strives to minimize its values. This minimax relationship between the two components facilitates the enhancement of the deblurring performance. The minmax game can be summarised as the following equation:

$$min_G max_D V(D,G) = E_{x \sim P_{data}} \log D(x) + E_z \log(1 - D(G(z))) \dots (1)$$

## 2    Generative Adversarial Networks Model

This section involves the design of the GAN model, including the architecture of the Generator and Discriminator, and as well as the loss functions. Table 1 shows the overall architecture of the GAN with 30 layers and 14.5M parameters in total.

|  | #Layers | K | S | $C_{in}$ | $C_{out}$ | #Parameters |
|---|---|---|---|---|---|---|
| Generator | 24 | 3 or 7 | 1 or 2 | 3 | 3 | 11399171 |
| Discriminator | 6 | 4 | 1 or 2 | 3 | 1 | 3098370 |
| Total | 30 |  |  |  |  | ≈14.5M |

Table 1: Summary of GAN architecture. K represents the kernel size, S represents the stride, $C_{in}$ is the input channel, $C_{out}$ is the output channel, and number of parameters can be calculated via the formula below:

$$\#Parameters = C_{out} \times (C_{in} \times K^2 + 1) \ldots (2)$$

## 2.1 Generator Network

The Generator is based on ResNet [6] blocks with the aim of sharper image reproduction. Each ResNet block contains a convolution layer, batch normalization layer and ReLU as activation layer. 9 ResNet block were used in total for upsampling of blurred image. Table 2 shows the summary of Generator architecture with 24 layers and around 11.4M parameters in total.

| Generator | K | S | $C_{in}$ | $C_{out}$ | #Parameters |
|---|---|---|---|---|---|
| input_1 | 7 | 1 | 3 | 3 | 0 |
| conv2d | 7 | 1 | 3 | 64 | 9472 |
| conv2d_1 | 3 | 2 | 64 | 128 | 73856 |
| conv2d_2 | 3 | 2 | 128 | 256 | 295168 |
| conv2d_3 | 3 | 1 | 256 | 256 | 590080 |
| conv2d_4 | 3 | 1 | 256 | 256 | 590080 |
| conv2d_5 | 3 | 1 | 256 | 256 | 590080 |
| conv2d_6 | 3 | 1 | 256 | 256 | 590080 |
| conv2d_7 | 3 | 1 | 256 | 256 | 590080 |
| conv2d_8 | 3 | 1 | 256 | 256 | 590080 |
| conv2d_9 | 3 | 1 | 256 | 256 | 590080 |
| conv2d_10 | 3 | 1 | 256 | 256 | 590080 |
| conv2d_11 | 3 | 1 | 256 | 256 | 590080 |
| conv2d_12 | 3 | 1 | 256 | 256 | 590080 |
| conv2d_13 | 3 | 1 | 256 | 256 | 590080 |
| conv2d_14 | 3 | 1 | 256 | 256 | 590080 |
| conv2d_15 | 3 | 1 | 256 | 256 | 590080 |
| conv2d_16 | 3 | 1 | 256 | 256 | 590080 |
| conv2d_17 | 3 | 1 | 256 | 256 | 590080 |
| conv2d_18 | 3 | 1 | 256 | 256 | 590080 |
| conv2d_19 | 3 | 1 | 256 | 256 | 590080 |
| conv2d_20 | 3 | 1 | 256 | 256 | 590080 |
| conv2d_21 | 3 | 1 | 256 | 128 | 295040 |
| conv2d_22 | 3 | 1 | 128 | 64 | 73792 |
| conv2d_23 | 7 | 1 | 64 | 3 | 9411 |

Table 2: Generator architecture

## 2.2 Discriminator Network

The Discriminator is to distinguish whether the receiving image is real or fake and output single value decision result, either 0(for fake image) or 1(for real image). LeakReLU, Sigmoid and Tanh functions were used for activation. Table 3 shows the summary of Discriminator architecture with 6 layers and around 3M parameters in total.

| Discriminator | K | S | $C_{in}$ | $C_{out}$ | #Parameters |
|---|---|---|---|---|---|
| input_1 | 4 | 2 | 3 | 3 | 0 |
| conv2d_24 | 4 | 2 | 3 | 64 | 3136 |
| conv2d_25 | 4 | 2 | 64 | 64 | 65600 |
| conv2d_26 | 4 | 2 | 64 | 128 | 131200 |
| conv2d_27 | 4 | 2 | 128 | 256 | 524544 |
| conv2d_28 | 4 | 1 | 256 | 512 | 2097664 |
| conv2d_29 | 4 | 1 | 512 | 1 | 8193 |

Table 3: Discriminator architecture

## 2.3 Loss Functions

Two loss functions were involved: Perceptual Loss at the end of the Generator and Wasserstein Loss at the end of the whole GAN.

2.3.1 Perceptual Loss [7]
   To ensure the GAN model is deblurring the images, perceptual loss was calculated directly on the output of the Generator and compared to first convolutions of VGG16 [8].

2.3.2 Wasserstein Loss [9]
   To improve the convergence of GAN, wasserstein loss was calculated on the output of the whole model by taking the mean difference of the two images.

## 3 Experiment and Result Analysis

This section involves the training and testing of the GAN model and performs in-depth analyzation on the results. The working environment for this project was on MacOS with 1.4 GHz Quad-Core Intel Core i5. It leverages the Keras library to implement the GAN model.

### 3.1 Dataset

GoPRO dataset [3] (2.12GB, light-weight version) was used to train the model. It provides a pair of clear and blur images from street views with around 500 images of size 1280x720. Note that the input of Generator will intake the blur image directly instead of noise based on Figure 1 since it helps to save the implementation time for constructing the Generator model.

### 3.2 Training

The training parameters are stated as below:
- Batch size: 16
- Training epochs: 40
- Learning rate: 1E-4
- Beta_1: 0.9
- Beta_2: 0.999

- Epsilon: 1E-08

The training process begins by dividing the input image into a 256x256 matrix and loading it into the model. The training then proceeds with the number of epochs listed as above, during which the data is divided into batches. The Generator and Discriminator are both trained using the two loss functions described earlier. Initially, the Generator generates a fake image, and the Discriminator is trained to distinguish between the fake image and real inputs. This process is repeated iteratively to train the entire GAN model. The training continues until all the input images have been used. The general training flow can be referred to Figure 1 and the training time loops for around 3 hours.

**3.3 Validation**

The performance of image deblurring is shown in Figure 2. The output image from GAN model has sharper pixels than before, achieving the image deblurring task. For example, the edges of the human in the first and second image in Figure 2 are sharper and clearer, forming a more convincing image.

Note that this GAN model is specifically designed to deblur motion-blurred images. It is trained on a dataset of motion-blurred images and corresponding sharp images to learn the mapping between the two. The model is capable of estimating the motion blur kernel and then using it to deblur the image. Hence, it may not be effective for deblurring images that are blurred due to enlargement, such as the third image showing the whiteboard of a classroom from Figure 2. As a result, the deburred performance is not that good. When an image is enlarged, the pixels are stretched and interpolated to create a larger image. This interpolation process can introduce blurring and loss of details. On the other hand, motion blur occurs when there is relative motion between the camera and the scene during the exposure time. This results in streaks or smearing of the image in the direction of the motion. As a result, the GAN's training and architecture are not optimized to handle it.

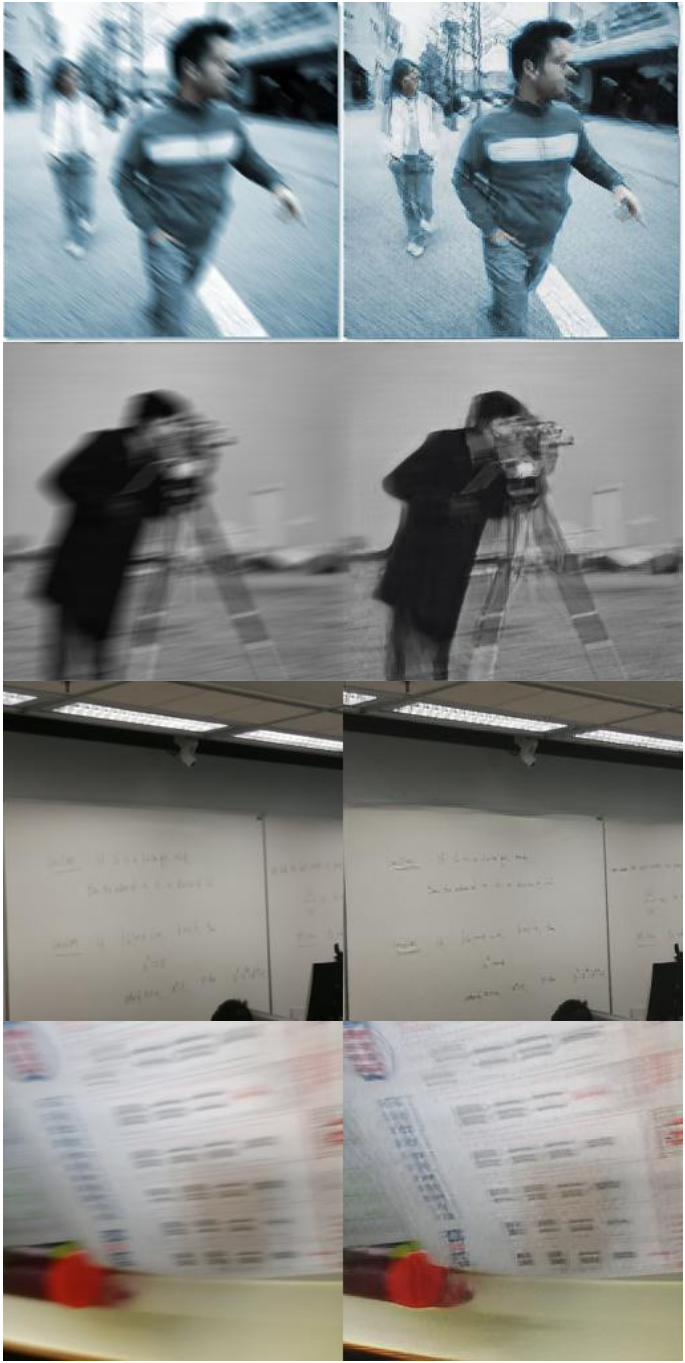

Figure 2: Image deblurring results (left images: input image of GAN, right images: deblurred image from GAN)

Two metrics, PSNR and SSIM were used to evaluate the deblurring performance. PSNR measures the ratio between the maximum possible power of a signal and the power of corrupting noise that affects the fidelity of the signal. It is calculated on the logarithm of the ratio of maximum possible pixel value to mean squared error (MSE) between the original and processed images. A higher PSNR value indicates a higher quality reconstruction. SSIM measures the structural similarity between the original and processed images. It takes into account three components: luminance, contrast, and structure respectively. SSIM is calculated by comparing the similarity of the above three components between corresponding patches of the original and processed images. The resulting SSIM value ranges from -1 to 1, with 1 indicating perfect similarity.

The summary of PSNR and SSIM of this project are shown on Table 4. This project achieves an average of 29.3 PSNR and 0.72 SSIM. Table 5 shows the comparison of the evaluation metrics with related works discussed in the Introduction Section. Compared to [4] which used 12 ResNet block in total, this project has simpler network architecture with 9 ResNet block only. Thus, the performance is not that good with respect to [4], but it is still better than [2] and [3] in terms of PSNR. However, this project provides a detail analyse on the number of layers and parameters on the GAN architecture, which are rarely discussed in the related works before. In the future, this project believes that the values of PSNR and SSIM can be further improved by increasing the training epochs and number of training images. These will help to optimise the performance of image deblurring using the GAN model and better than the other related works.

| Evaluation metric | Maximum value | Minimum value | Mean value |
| --- | --- | --- | --- |
| PSNR | 32.33 | 25.39 | 29.30 |
| SSIM | 0.77 | 0.65 | 0.72 |

Table 4: Evaluation based on the PSNR and SSIM

| Method | PSNR | SSIM |
| --- | --- | --- |
| Kupyn *et al* [2] | 28.7 | 0.958 |
| Nah *et al* [3] | 28.93 | 0.91 |
| Ji *et al* [4] | 31.15 | 0.96 |
| This work | 29.3 | 0.72 |

Table 5: Comparison on PSNR and SSIM with related works

## 4	Conclusion

In conclusion, this project focused on utilizing Generative Adversarial Networks (GAN) as a solution for image deblurring in computer vision. The proposed approach involved defining a GAN model in TensorFlow and training it with the GoPRO dataset. The Generator in the GAN model took in blurry images and generated fake images to convince the Discriminator, which received clear images and distinguished between real and fake images. After that, the pretrained GAN model was used to deblur motion-blurred images from daily life as well as a testing set for validation. The results showed that the pretrained GAN network was able to produce sharper pixels in images, achieving an average of 29.3 PSNR and 0.72 SSIM. This approach effectively addressed the challenges posed by image blurring and led to the generation of visually pleasing and sharp images. By leveraging the adversarial learning framework of GAN, the proposed approach has the potential for real-world applications in image restoration. It offers a promising solution for generating clearer images from blurry inputs caused by factors such as motion blur. The use of GAN in image deblurring opens up possibilities for improving the quality of images in various domains, including but not limited to photography, surveillance, and medical imaging, etc.